%% file: ms.tex
\documentclass[journal]{IEEEtran}

\usepackage{url}
\usepackage{booktabs}
\usepackage{graphicx}
\usepackage{float}
\usepackage{flushend}
\usepackage[table]{xcolor}
\usepackage{amsmath}
\usepackage{amssymb}
\usepackage{subcaption}

\usepackage{array,setspace, amsfonts,url,bm, cellspace}
\usepackage{tikz}
\usepackage{epstopdf}
\usetikzlibrary{positioning}
\usetikzlibrary{arrows}
\usepackage{pifont}
\usepackage{multirow, hhline}
\usetikzlibrary{decorations.pathreplacing,angles,quotes}
\usepackage[flushleft]{threeparttable}

\begin{document}

\title{REGINA - Reasoning Graph Convolutional Networks in Human Action Recognition}

\author{Bruno Degardin, Vasco Lopes and Hugo Proen\c{c}a,~\IEEEmembership{Senior Member,~IEEE} \\

\thanks{B. Degardin and H. Proen\c{c}a are  with the IT: Instituto de Telecomunica\c{c}ões, Department of Computer Science, University of Beira Interior, Covilha, Portugal. V. Lopes is with the NOVA LINCS, University of Beira Interior, Portugal. E-mail: \{bruno.degardin, vasco.lopes\}@ubi.pt, hugomcp@di.ubi.pt}}

\markboth{ArXiv}%
{Shell \MakeLowercase{\textit{et al.}}: Bare Demo of IEEEtran.cls for IEEE Journals}
%



\def\etal{\emph{et al}.}

\maketitle

\begin{abstract}
It is known that the kinematics of the human body skeleton reveals valuable information in action recognition. Recently, modeling skeletons as spatio-temporal graphs with Graph Convolutional Networks (GCNs) has been reported to solidly advance the state-of-the-art performance. However, GCN-based approaches exclusively learn from raw skeleton data, and are expected to extract the inherent structural information on their own. This paper describes \textit{REGINA}, introducing a novel way to \textit{RE}asoning \textit{G}raph convolutional networks \textit{IN} Human \textit{A}ction recognition. The rationale is to provide to the GCNs additional knowledge about the skeleton data, obtained by handcrafted features, in order to facilitate the learning process, while guaranteeing that it remains fully trainable in an end-to-end manner. The challenge is to capture complementary information over the dynamics between consecutive frames, which is the key information extracted by state-of-the-art GCN techniques. Moreover, the proposed strategy can be easily integrated in the existing GCN-based methods, which we also regard positively. Our experiments were carried out in well knwon action recognition datasets and enabled to conclude that REGINA contributes for solid improvements in performance when incorporated to other GCN-based approaches, without any other adjustment regarding the original method. For reproducibility, the REGINA code and all the experiments carried out will be publicly available at \url{https://github.com/DegardinBruno}.
\end{abstract}


%
\IEEEpeerreviewmaketitle

\section{Introduction}
\label{sec:introduction}
The remarkable effectiveness of the human visual perception system to recognize motion patterns has been inspiring decades of research in the computer vision community \cite{cheng2020skeleton, qiu2017learning, shi2019two, tran2015learning, tran2018closer, yan2018spatial}. In particular, the ability to instinctively recognize other humans actions has been vital for the survival of the specie itself. However, action recognition can be particularly hard to perform automatically, due to the sophistication and complexity of movements, the inter-dissimilarity between humans, the background clutter and viewpoints \cite{aggarwal2011human, poppe2010survey, weinland2011survey}. This is an extremely crowded topic in the research community, where skeleton-based approaches have been gaining popularity, due to their ability to partially handle these challenges: they are known to faithfully handle variations in appearance and cluttered backgrounds, which allowed the pioneering skeleton-based action recognition techniques \cite{junejo2010view, ohn2013joint, vemulapalli2014human, vemulapalli2016rolling} to learn and extract relevant patterns.

\begin{figure}
  \begin{center}
  \includegraphics[width=8.85cm]{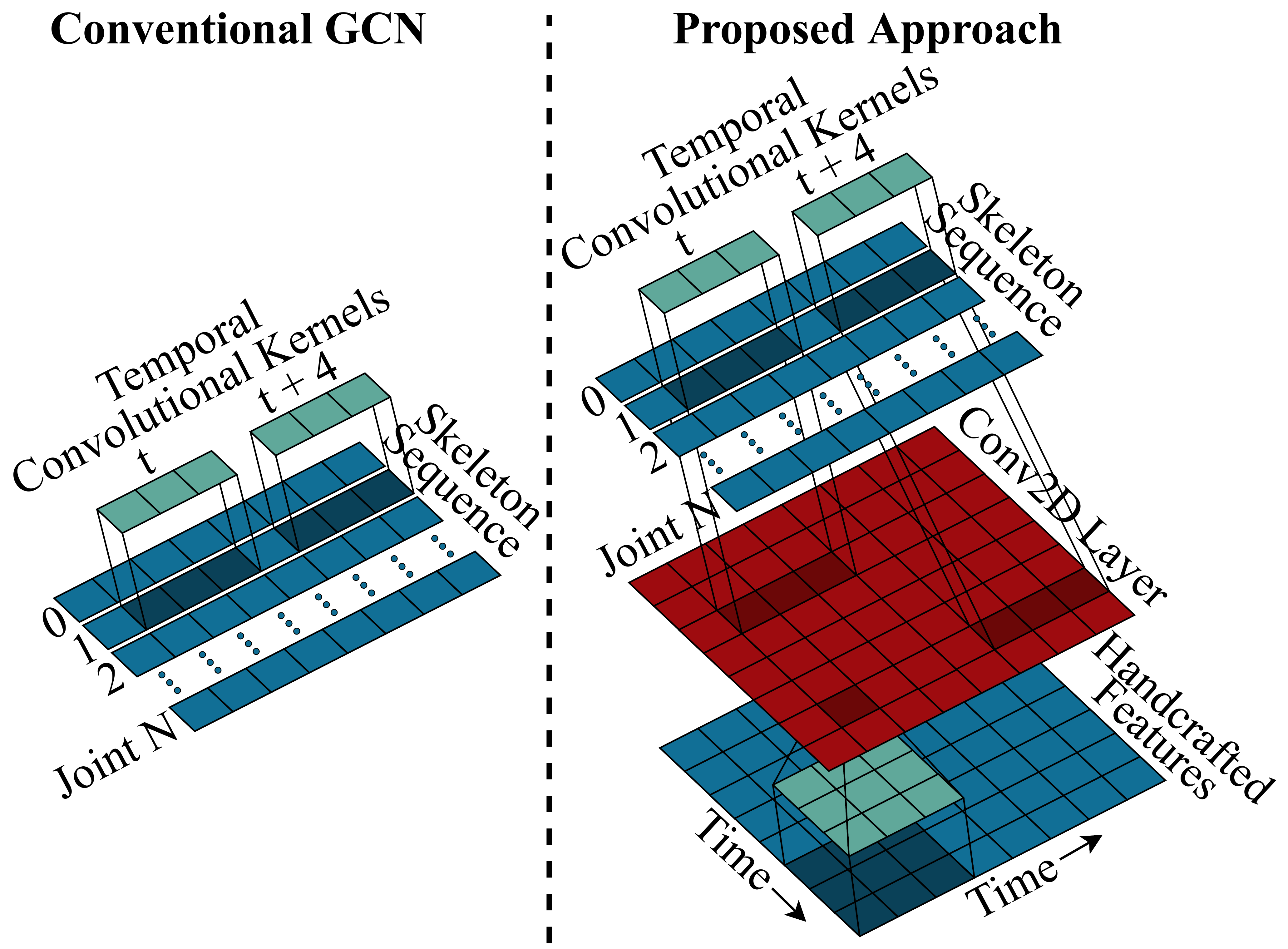}
  \caption{Rationale for the strategy proposed to \textbf{improve temporal graph convolution} strategies. When compared  to the conventional scheme (at left), we consider two additional components: 1) a set of handcrafted features (lower layer on the right side); that are integrated into the temporal convolutional kernel patches (light blue), according to 2) a learnable temporal convolutional layer (in red). This information is further fused to the skeleton sequence information, according to the conventional temporal convolutional kernels.}
  \label{fig:regina_unit}
  \end{center}
\end{figure}

The appearance of data-driven approaches (i.e., deep learning-based)  facilitated the perception of human poses on image data, through efficient pose estimation algorithms \cite{cao2019openpose, cao2017realtime, Neverova2019DensePoseConfidences, zhang2012microsoft}, which has propelled a growing number of skeleton-based techniques and large-scale datasets \cite{kay2017kinetics, liu2019ntu, shahroudy2016ntu}. This ability to extract high-level features with deep learning-based architectures has paved the way to structured skeleton-based approaches, such as convolutional neural networks (CNNs) with pseudo-images \cite{ke2017new, liu2017enhanced, soo2017interpretable} and recurrent neural networks (RNNs) with sequence coordinate vectors \cite{liu2016spatio, song2017end, zhang2017view}. More recently, graph convolutional networks (GCNs) were introduced \cite{kipf2016semi}, to model skeleton data by means of spatiotemporal graph convolutions \cite{yan2018spatial}. Due to its intrinsic ability to exploit the skeleton inherent graph structure, several variants were subsequently introduced, typically proposing additional modules \cite{li2019spatio, shi2019skeleton, si2019attention, zhang2020semantics} and multi-streams \cite{cheng2020skeleton, liu2020disentangling, shi2019two} strategies.

\begin{figure*}[t]
  \begin{center}
  \includegraphics[width=\textwidth]{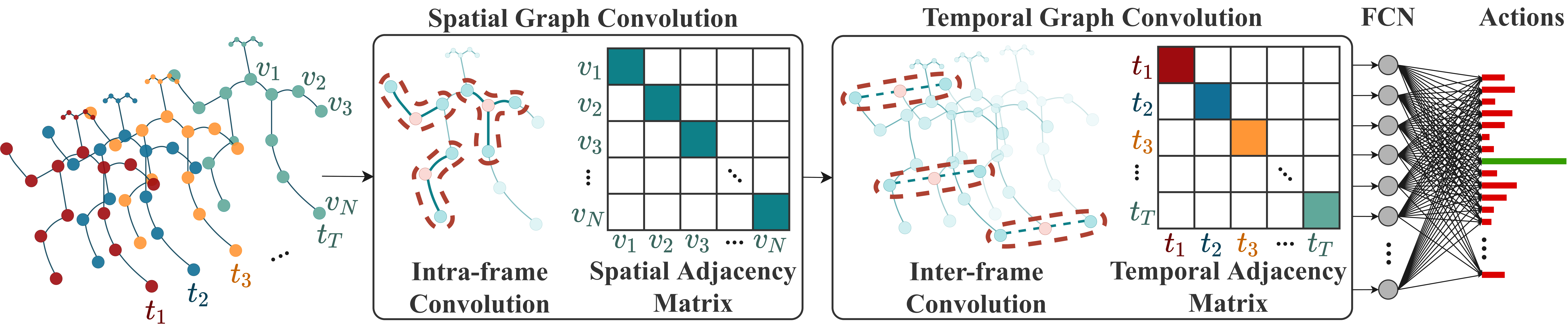}
  \caption{\textbf{Cohesive view of the REGINA incorporation in a spatio-temporal graph convolutional network (ST-GCN)} \cite{yan2018spatial}. A skeleton sequence is used as input of the network, where a spatio-temporal graph is constructed over the sequence. The spatial adjacency matrix handles the intra-frame (spatial) convolutions, while our temporal adjacency matrix handles the inter-frame (temporal) convolutions. The temporal graph convolution will consider the skeleton joint information (from the spatial graph convolution) and its global information in terms of temporal progression through the optimized temporal adjacency matrix. Finally, a standard Softmax layer predicts the corresponding action class.}
  \label{fig:regina_overview}
  \end{center}
\end{figure*}

In spite of the rapid progresses observed in recent years, two major shortcomings remain: 1) the high-level formulation of temporal graph convolutions constrains its design in computing them over a root node temporal neighbor set. Despite its neighborhood importance, this results in losing the skeleton global view, in particular concerning its temporal evolution; and 2) current methods use a spatial adjacency matrix to define the graph spatial topology and only assume its natural temporal adjacency, pre-defining the temporal receptive fields heuristically. Even though \cite{cheng2020skeleton, shi2019skeleton, shi2019two} present a learnable spatial adjacent matrix and graph shifting, our experiments point for local and temporal limitations of the encoding process, losing the holistic perspective in their temporal convolutions.

As a response to the above problems, in this paper we describe REGINA, a novel graph convolution architecture that can be easily integrated into the existing GCN-based techniques and enables solids improvements in performance with respect to the baselines. REGINA extends GCNs by employing shallow temporal representations to obtain a learnable temporal adjacency configuration, adjusting the temporal receptive fields. The proposed architecture is composed of a new graph convolution layer that learns the temporal importance of actions, which is trainable in an end-to-end manner, exploiting the skeleton data in a complementary way apart motion between frames.

The proposed strategy addresses naturally the previously mentioned problems: instead of using GCNs that only consider the temporal adjacency, we integrate a handcrafted self-similarity matrix (SSM) to the temporal graph convolution part, aimed to enhance the global connectivity across the temporal axis. Specifically, by extracting the SSM with point-wise convolutional kernel patches, the temporal graph receptive fields can be weighted based in the skeleton corresponding temporal differences, assigning local weights in direct proportion to the difference values and, consequently, considering an action \emph{holistic view}. As illustrated in Fig. \ref{fig:regina_unit}, the SSM is fed into a 2-dimensional convolutional layer before weighting each sample from the temporal convolution kernel. As different graph convolutional layers may require different temporal receptive fields \cite{xie2018rethinking, li2019collaborative}, the temporal configuration is learned and adjusted based on the SSM descriptor. The self-similarity descriptor choice is justified empirically, based in the experiments carried out, which showed the robustness of the SSM descriptor to pose and camera variance. Additionally, as described in Section \ref{sec:integration}, REGINA outperforms the conventional temporal convolutions strategies and can be easily integrated to state-of-the-art GCN-based techniques, enabling consistent improvements in performance in well known datasets. A cohesive view of the proposed REGINA strategy, incorporated within a spatiotemporal graph convolutional network is given in Fig.~\ref{fig:regina_overview}.

In summary, our main contributions are three-fold: 1) We propose a graph convolution architecture that fuses shallow temporal representations to spatiotemporal GCNs, enabling to retain the skeleton global information, while extracting other higher-level spatiotemporal semantic features; 
2) we experimentally show the ease of incorporating REGINA into the state-of-the-art GCN-based approaches, which does not prevent them from being trained in an end-to-end manner, enabling solid improvements in performance over the baselines;
3) we report competitive results with respect to the state-of-the-art techniques in a large-scale dataset (NTU RGB+D) for skeleton-based action recognition.\\

The remainder of this paper is organized as follows: Section~\ref{sec:related} summarizes the related work and in Section~\ref{sec:REGINA} we describe in detail the REGINA strategy. Section~\ref{sec:Results} discusses our experiments and results and, finally, Section~\ref{sec:Conclusions} concludes this paper.

\section{Related Work}
\label{sec:related}
Due to the rise of deep learning-based architectures over the last decade, an evolution in the action recognition challenges has been witnessed. From the temporal information extraction breakthrough in video data, using 3-dimensional convolutional networks (3D ConvNets) \cite{qiu2017learning, tran2015learning, tran2018closer}, to the successful application of graph convolutional networks (GCNs) \cite{cheng2020skeleton, shi2019two, yan2018spatial} to model skeleton-based data as a graph over the spatial and temporal domains.

\subsection{Skeleton-based Action Recognition}
\label{ssec:skeleton}
Human pose estimation \cite{guler2018densepose, xiu2018pose, zhang2012microsoft} is acknowledged as one of the most important cues in behavior analysis/action recognition topics. Typically, pose estimation techniques provide 3D representations of the subject skeleton, attenuating the variations in appearance that RGB and depth data representations contain, while obtaining semantically rich and very descriptive representations that drive the learning process exclusively over human dynamics. Consequently, skeleton data ignited a variety of scientific works in action recognition, from the pioneering techniques based in handcrafted features to the current data-driven state-of-the-art approaches (deep learning-based). 

Handcrafted feature methods capture the actions dynamics patterns by designing algorithms based on physical human foreknowledge. Such patterns can be self-similarity descriptors \cite{junejo2010view} to capture the temporal evolution of actions in a nearly view-independent paradigm, or Lie group curves \cite{vemulapalli2014human, vemulapalli2016rolling}, to model the relative geometry between different body parts (providing a more meaningful representation than their absolute locations), or even covariance matrixes of joint trajectories \cite{hussein2013human}, that hierarchically encode the temporal dependencies.

The existing deep learning-based architectures can be further classified into three categories. CNN-based methods, that represent the skeleton sequence data as pseudo-images through pre-processing transformations, such as temporal cylindrical coordinates\cite{ke2017new}, color series representing the skeletons \cite{liu2017enhanced}, or temporal concatenation of the skeleton sequence raw coordinates \cite{soo2017interpretable}. RNN-based approaches, which are commonly applied over each human body joint coordinate vector, model the skeleton data as a sequence to learn its long-term contextual information, with a focus on resorting to LSTMs  \cite{liu2016spatio, song2017end, zhang2017view}. More recently, spatiotemporal GCNs successfully generalized the convolution from image to graph \cite{cheng2020skeleton, shi2019two, li2019actional, tang2018deep, yan2018spatial}, being justified by their improved representation of structural information embedded in skeleton data, modeling the data as a graph with vertices (joints) and edges (bones), rather than vector sequences or 2D grids. This kind of models has been consistently advancing the state-of-the-art performance over the last years.

\subsection{Graph Convolutional Networks}
\label{ssec:gcns}

Considering that skeleton information can be naturally embedded in graphs, the work by Yan \etal~\cite{yan2018spatial} inspired several variants of spatiotemporal GCNs, by modeling skeletons as graph structures. The basis is that, for each node (joint), a layer-wise update is performed with a heuristically predefined neighborhood (adjacent joints). Additional modules were proposed, such as graph directions \cite{shi2019skeleton} and attention mechanisms \cite{si2019attention}. Observing that the human body physical connections tend to constraint the learned graph structure, Shi \etal~\cite{shi2019two} proposed an adaptive adjacency matrix. More recently, spatial and temporal shift convolutions were also considered \cite{cheng2020skeleton}, having as goal to increase the flexibility of the resulting graph structure.

Given the important role of GCNs in skeleton-based action recognition, we observed that most state-of-the-art approaches underestimate the importance of handcrafted-based features, due to the challenge in representing the whole human skeleton dynamic complexity. Notwithstanding, we consider that handcrafted-based techniques  still provide valuable information about the skeleton spatial arrangement and temporal kinematics. In a way similar to the well-known optical flow technique \cite{horn1981determining}, there is no doubt that CNN-based methods significantly overcome the performance of optical flow-based methods. However, fusion optical-flow to CNN-based features  \cite{simonyan2014two, wang2015action, wang2016temporal, zhang2016real} not only advanced the state-of-the-art but has also inspired future works. Our proposal, REGINA, is the first to use handcrafted features at the core of GCNs. In opposition to previously published approaches, it is the first of its kind to provide an easy integration to other methods, by not modeling the raw skeleton information from scratch, and - instead - providing foreknowledge to GCNs as a way to conduct their learning process.

\section{REGINA: Proposed Graph Convolutional Architecture}
\label{sec:REGINA}
As stated above, our rationale is that the richness of skeleton-based data expedites the understanding of human actions. In this section, we describe in detail our proposed graph convolutions architecture that fuses skeleton-based shallow representations into graph convolutional units, which enables to analyze complementary information and enhances the effectiveness of GCNs on this task.

\subsection{Reasoning Skeleton Information}
\label{ssec:reason}
The pioneering action recognition techniques were able to extract consistent patterns of the human kinematics by resorting to foreknowledge formulations that analyse the natural body dynamics and extract skeleton information.

Having noticed the constrained structure of typical GCN-based techniques due to their high-level formulation, we propose to consider self-similarity descriptors between representations of time-frame pairs. Such kind of information, combined with skeleton-based sequences, explicitly describes the temporal difference evolution for each time frame skeleton pose, which provides additional information to the networks that should useful in the optimization process (learning). The underlying idea is that - this way - the graph is not required anymore to implicitly estimate the global skeleton information, and can analyze other types of (orthogonal) features, making the recognition of human actions an easier task.

\begin{figure}[t]
  \begin{center}
  \includegraphics[width=8.4cm]{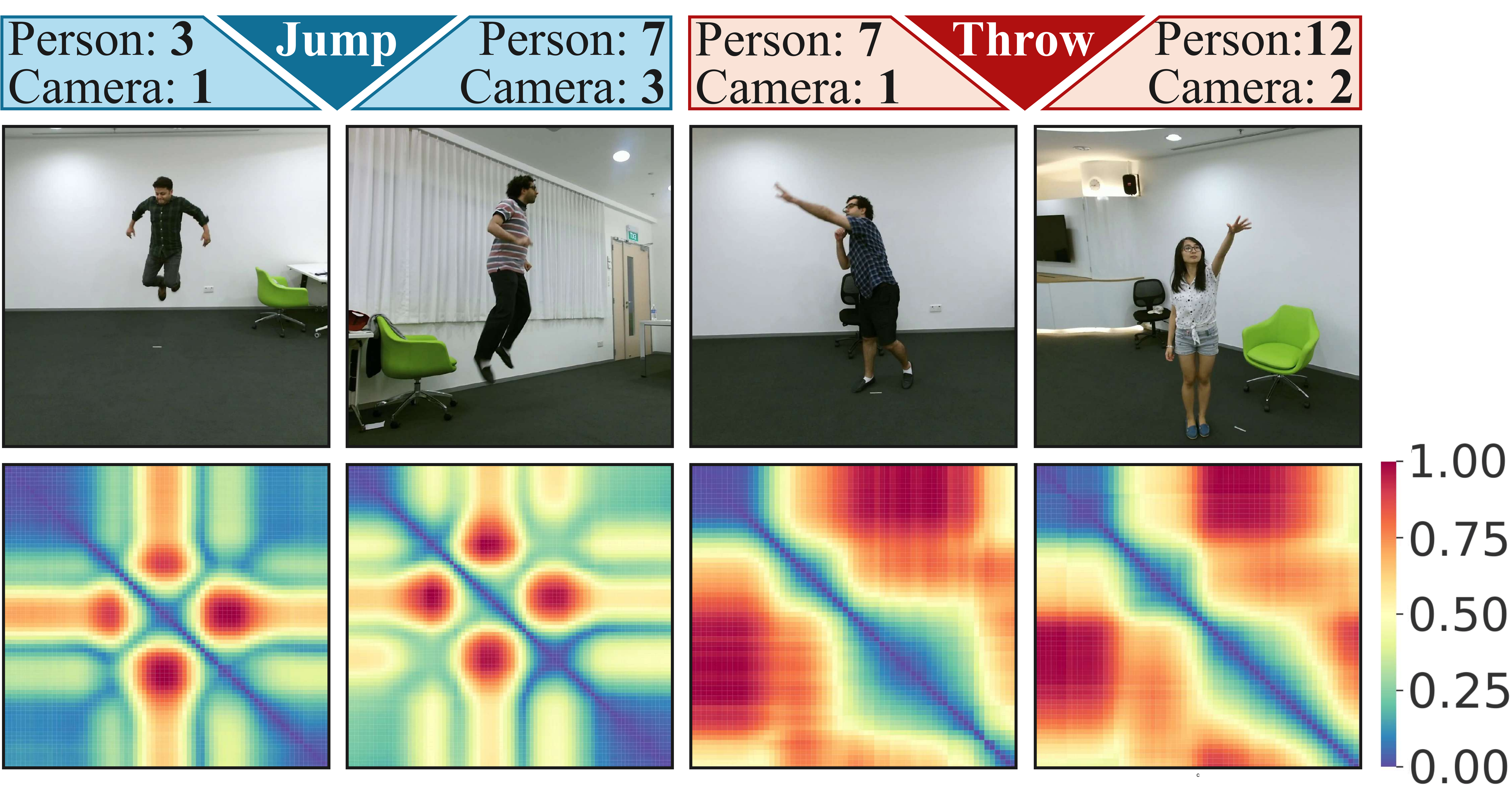}
    \caption{\textbf{Example of the Self-Similarity Matrix (SSM) intra and inter-action variance over the NTU RGB-D} \cite{shahroudy2016ntu} dataset. The first-row illustrates two actions done by different people and under different pose (\textit{Jump Up}, in the leftmost columns and \textit{Throw}, in the rightmost columns). The bottom row provides the corresponding SSM yielding from the corresponding skeleton sequences.}
  \label{fig:ssm_comparison}
  \end{center}
\end{figure}

Having observed the SSM features permanence with respect to variations in pose and identity, we decided to use this kind of handcrafted features in our action recognition experiments. However, it should be noted that the proposed strategy can use other types of handcrafted information, that can seemingly integrated in our architecture without any other adaptions required. However, for the proof-of-concept reported in this paper, we resorted to the use of SSM in all experiments. As illustrated in Fig. \ref{fig:ssm_comparison}, the SSM patterns can easily provide discriminating/permanent auxiliary information to the GCN, enabling the network to analyze other kinds of information. In this example, regarding the action "jump up", the two \textit{hotspots} (e.g., in lower diagonal) reflect the body's movement's flexion (before jumping) and the highest attainable height from jumping up, the differences in the resulting SSM values are obvious.

\begin{figure}[h]
  \begin{center}
  \includegraphics[width=8.4cm]{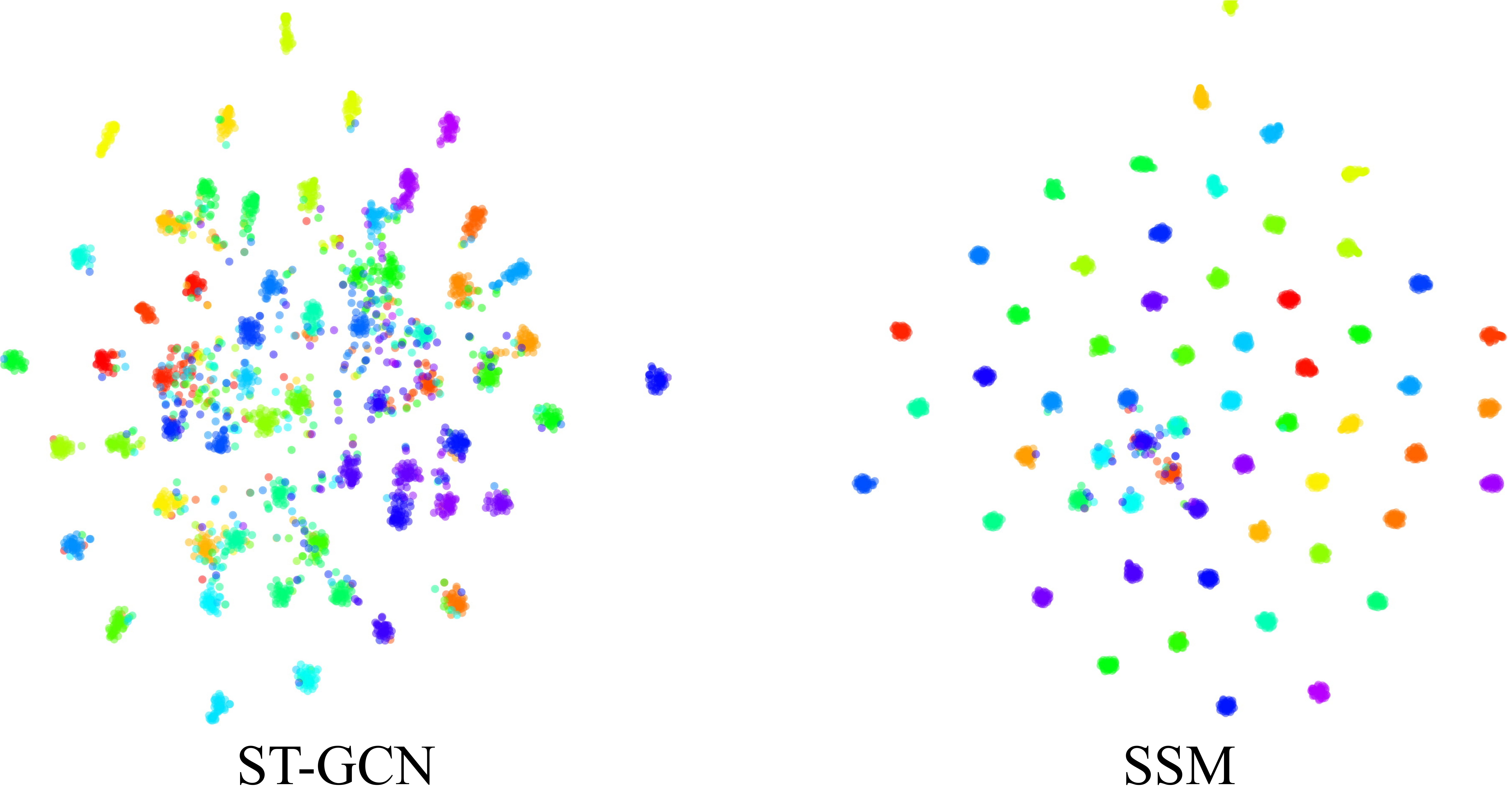}
    \caption{\textbf{Feature Embedding}. Illustration of the Feature embedding (using t-SNE \cite{van2008visualizing}) obtained for ST-GCN \cite{yan2018spatial} and SSM features on the cross-view validation benchmark of the NTU RGB+D \cite{shahroudy2016ntu} dataset. Each point represents one skeleton video sequence, with elements belonging to the same class (total of 60 classes) appearing in the same color.}
  \label{fig:feat_embed}
  \end{center}
\end{figure}

Aiming at perceiving the robustness of SSM descriptor as  generic features, we qualitatively evaluated them by visualizing the feature embedding yielding from a state-of-the-art action recognition technique (ST-GCN)and SSM. We randomly selected 50 skeleton sequences of each class from the cross-view validation benchmark of the NTU RGB+D \cite{shahroudy2016ntu} dataset, and then extracted the features of the skeleton sequences using our primary baseline pre-trained on the respective training benchmark, spatiotemporal graph convolutional networks (ST-GCN) \cite{yan2018spatial}, and the SSM descriptor used in our experiments. These features were then projected into the 2-dimensional space using t-SNE \cite{van2008visualizing} technique. Figure \ref{fig:feat_embed} provides an illustration of the resulting embedding, where colors represent the different actions (classes), and each point represents one instance of the corresponding class.

Formally, let $\boldsymbol{\mathcal{V}} \in \mathbb{R}^{n \times t \times c}$ denote the skeleton sequence representation with $n$ joints, $t$ frames, and $c$ dimensions of joint coordinates. Given a set of skeleton points $\mathcal{V}_j = \{v_{ij}\}$, $ i \in  \{1, \dots, n\}$ at time $t$, in this paper, we obtain the self-similarity matrix $\mathcal{S}$ using the mean Euclidean distance ($\ell-2$ norm) between skeleton poses at any moments $p$ and $q$ of the sequence: 

\begin{equation}\label{eq:1}
   d_{pq} = \frac{1}{N}\sum_{i} \|v_{ip} - v_{iq}\|_2,
\end{equation}
where $v_{ip}$ and $v_{iq}$ denote corresponding skeleton joints at times $p$ and $q$ on the sequence. Consequently, by obtaining a distance value between representations for all time-frame pairs, the self-similarity matrix descriptor yields a hollow square symmetric matrix of size $t\times t$.

\subsection{REGINA + Graph Convolutional Networks}
\label{ssec:graph}

The recent appearance of Graph Convolutional Networks (GCNs) enabled the extraction of embedded patterns over the spatial and temporal axes of a skeleton sequence, essentially by generalizing convolutions from images to graphs. However, due to their high-level formulation, graph convolutions are performed locally by taking into account exclusively a root node neighboring set. Recent studies reported that considering only local connections might be sub-optimal for recognizing skeleton-based actions \cite{cheng2020skeleton, li2019actional, shi2019skeleton, shi2019two}. These findings provided our motivations to take advantage of the valuable hard-coded knowledge from handcrafted shallow representations and design a graph convolution that not only considers an action holistic perspective in conventional graph convolutions, but can also be easily integrated into the current GCN-based approaches. 

Let the skeleton-based data used in GCN be denoted as a spatiotemporal graph $\mathcal{G} = (\mathcal{V}, \mathcal{E})$, with $n$ skeleton joints and $t$ frames. Therefore, the skeleton sequence's feature map can be denoted as $\textbf{X} \in \mathbb{R}^{n \times t \times c}$, with $c$ channels representing the joints coordinates. Typically, in spatial graph convolutions, one adjacency matrix $\textbf{A}$ and one identity matrix $\textbf{I}$ are used to define the intra-body joints connections, which can be dismantled into three partitions $p$ (regarding the set of neighbors resulting from the spatial configuration proposed in \cite{yan2018spatial}) where $\textbf{A} + \textbf{I} = \sum_p$ $\textbf{A}_p$. In a single frame, the graph convolution is given by:

\begin{equation}\label{eq:2}
   \textbf{Y} = \sum ^{p} _{i=1}\mathbf{\Lambda}_{i}^{-\frac{1}{2}}\textbf{A}_i\mathbf{\Lambda}_{i}^{-\frac{1}{2}}\textbf{X}\textbf{W}_i,
\end{equation}
where the degree matrix $\mathbf{\Lambda}_{p}^{ii} = \sum _j (\mathbf{A}_p^{ij})$ denotes the number of edges attached to each joint node, to normalize the adjacency matrix $\textbf{A}_p$. $\textbf{W}_p$ are the stacked weight vectors for each partition group $p$. 

Over the temporal axis, graphs are defined over consecutive frames, where most skeleton-based GCN approaches \cite{shi2019skeleton, shi2019two, yan2018spatial, zhang2020semantics} use 1-dimensional kernels as the conventional temporal graph convolution operators, where the receptive fields are defined beforehand. Despite the temporal neighborhood importance, it is obvious that the skeleton global perspective is lost in this operation.

Considering the above weaknesses, we propose an adaptive and learnable graph convolution network that mainstreams the temporal evolution expressiveness (from the skeleton holistic perspective) over the corresponding temporal receptive fields from the temporal graph convolutions, that can also be interpreted as a learnable temporal adjacency matrix. As stated above, the self-similarity matrix $\mathcal{S}$ explicitly describes the temporal difference for each time-frame with respect to every other time-frames. However, since multiple graph convolutional layers are used, different layers contain multilevel semantic information, and the raw integration of $\mathcal{S}$ would result in the same rigid and pre-defined temporal configuration to every layer. For our purposes, it was considered essential to adapt the handcrafted features to different semantic levels. Hence, we resorted to a convolutional layer over the self-similarity descriptor, in order to adaptively learn the optimized temporal configuration that fits optimally the hierarchical structure of GCNs. An illustration of this idea is provided in our reasoning graph convolutional unit in Fig. \ref{fig:regina_unit}. Hence, our major contribution can be defined as:

\begin{equation}\label{eq:3}
    \boldsymbol{\mathcal{R}_k} = \big((\boldsymbol{\mathcal{S}}_k\mathbf{W})^\top\boldsymbol{\mathcal{J}}_k\big)^\top,
\end{equation}
where $\boldsymbol{\mathcal{S}}_k$ is the corresponding temporal evolution of the SSM with respect to the temporal indices of kernel patch $k$, and $\boldsymbol{\mathcal{J}}_k$ is an all-ones vector of size $c$. This way, we are able to adaptively optimize our handcrafted descriptor, and the reasoning term is obtained for all feature channels in the graph convolution by incorporating \eqref{eq:3} in \eqref{eq:2}, according to the Hadamard product:

\begin{equation}\label{eq:4}
   \textbf{Y} = \sum ^{P} _{p=1}\mathbf{\Lambda}_{p}^{-\frac{1}{2}}\textbf{A}_p\mathbf{\Lambda}_{p}^{-\frac{1}{2}}\big(\textbf{X} \odot \boldsymbol{\mathcal{R}}\big)\textbf{W}_p
\end{equation}

By incorporating the 2D convolution matrix directly into the skeleton receptive fields of graph convolutions, we are able to provide global information of the corresponding skeleton without dismissing its partition strategies applied to the neighbor set of each node. Furthermore, due to its simple formulation, our contribution in \eqref{eq:3} is able to be easily integrated into other GCN-based methods, without preventing them from being trained in an end-to-end manner, as described in Section \ref{sec:integration}.

\section{Experiments and Discussion}
\label{sec:Results}
In this section, we start by comparing the effectiveness of two state-of-the-art GCN-based techniques with/without our contribution (REGINA). Then, we compare our best performing models to the state-of-the-art approaches over both benchmarks of the well known NTU RGB+D \cite{shahroudy2016ntu} dataset.

\subsection{Dataset, Evaluation metrics and Experimental settings}

\textbf{Dataset}. Our experiments were conducted in the well-known NTU RGB-D \cite{shahroudy2016ntu} dataset, which is currently the most widely used set for evaluating skeleton-based action recognition methods. This set is composed of 56,880 video samples, with 60 action classes. Data were captured from highly restricted camera views providing 3D skeleton and RGB-D data from 40 volunteers for each action sample, with 25 joints for each skeleton. The authors have recommended two benchmarks: 1) cross-subject, where models are trained with 20 subjects and tested with the remaining 20 subjects; and the 2) cross-view setting: where models are trained with camera views 2 and 3 and tested on camera view 1.

\textbf{Evaluation metrics}. We followed the NTU RGB-D \cite{shahroudy2016ntu} convention and evaluated the recognition performance by reporting the Top-1 and Top-5 recognition accuracy on both benchmarks.

\textbf{Experimental settings}. The skeleton sequences temporal length were normalized to $t = 300$ frames. Therefore, each self-similarity descriptor corresponding to one skeleton sequence is defined by a $300\times 300$ matrix. For a fair comparison, the skeleton data was pre-processed as described in \cite{yan2018spatial} and \cite{shi2019two} for the ST-GCN and 2s-AGCN experiments, respectively. The methods that were used as baseline comparisons with/without the proposed REGINA were not adapted in any manner. The unique adaptation was that, when considering our proposal,  instead of the one-dimensional graph convolutions, we used our temporal graph convolution scheme. All experiments reported in this paper follow these guidelines and settings.

\subsection{Baselines Comparison}
\label{sec:integration}

We evaluated the REGINA effectiveness with respect to two state-of-the-art GCN approaches over both evaluation protocols from the large-scale benchmarks, NTU RGB-D \cite{shahroudy2016ntu} dataset. Furthermore, we combined our approach with multi-stream architectures to illustrate the robustness of our proposal to different kinds of features.

\def\@fnsymbol#1{\ensuremath{\ifcase#1\or *\or \dagger\or \ddagger\or
   \mathsection\or \mathparagraph\or \|\or **\or \dagger\dagger
   \or \ddagger\ddagger \else\@ctrerr\fi}}
\newcommand{\ssymbol}[1]{^{\@fnsymbol{#1}}}


\newcolumntype{C}[1]{>{\centering\arraybackslash}p{#1}}
\newcolumntype{l}[1]{>{\arraybackslash}p{#1}}

\begin{table}[h]
\begin{center}
\begin{tabular}{|C{0.001\textwidth}|l{0.15\textwidth}|C{0.044\textwidth}|C{0.044\textwidth}|C{0.044\textwidth}|C{0.044\textwidth}|}
\cline{2-6}
\multicolumn{1}{c|}{} & 
\textbf{\footnotesize{Method}} & 
\textbf{\footnotesize{$\mathbf{Top\text{-}1}$}} &  \textbf{\footnotesize{$\mathbf{Top\text{-}5}$}}& \textbf{\footnotesize{$\mathbf{Top\text{-}1}$}} &  \textbf{\footnotesize{$\mathbf{Top\text{-}5}$}} \\
\cline{2-6}
\noalign{\smallskip}
\hline
\parbox[t]{1mm}{\multirow{5}{*}{\hspace{-0.21em}\rotatebox[origin=c]{90}{\footnotesize{\textbf{NTU RGB-D}}}}}
& \multicolumn{1}{c}{ \cellcolor{gray!25} }
& \multicolumn{2}{c}{ \cellcolor{gray!25}  \footnotesize{\emph{Cross-Subject} }}
& \multicolumn{2}{c|}{ \cellcolor{gray!25}  \footnotesize{\emph{Cross-View} }}\\
&\scriptsize{ST-GCN~\cite{yan2018spatial}} & 
\scriptsize{$81.51$}  & 
\scriptsize{$96.92$}  &
\scriptsize{$88.34$}  & 
\scriptsize{$98.33$}  \\
&\scriptsize{\textbf{ST-GCN} + \textbf{REGINA}} & 
\scriptsize{$\mathbf{83.62}$}  & 
\scriptsize{$\mathbf{97.71}$}  & 
\scriptsize{$\mathbf{90.12}$}  & 
\scriptsize{$\mathbf{99.00}$}  \\
\cline{2-6}
&\scriptsize{2s-AGCN~\cite{shi2019two}} & 
\scriptsize{$\mathbf{88.51}$}  & 
\scriptsize{$\mathbf{98.43}$}  &
\scriptsize{$95.12$}  & 
\scriptsize{$99.14$}  \\
&\scriptsize{\textbf{2s-AGCN} + \textbf{REGINA}} & 
\scriptsize{$88.44$}  & 
\scriptsize{$98.22$}  &
\scriptsize{$\mathbf{95.21}$}  & 
\scriptsize{$\mathbf{99.38}$}  \\\hline

\end{tabular}
\vspace{-1em}
\end{center}
\caption{Comparison of \textbf{REGINA + GCN-based results}. Performance comparison when fusing the proposed REGINA architectures to two state-of-the-art methods. Results are provided  for both benchmarks of the NTU RGB+D \cite{shahroudy2016ntu} set.}
\label{tab:ablation_full}
\end{table}

\textbf{GCN-based + REGINA fusion}. In order to verify REGINA effectiveness when fused to GCN-based techniques, we considered two baselines: the spatiotemporal GCN \cite{yan2018spatial}, ST-GCN, and a two-stream adaptive GCN (2s-AGCN) \cite{shi2019two}, being both described in Section \ref{sec:related}. As these baselines use one-dimensional kernels in their temporal convolutions, some adaptations were carried out in the original implementations to add our temporal graph convolutions scheme in each method, without any further changes in each original framework. Noting that REGINA does not prevent any of those methods from being trained in an end-to-end manner, we also used the default learning settings,  reporting the obtained performance in Table \ref{tab:ablation_full}. Results show that the REGINA scheme was able to improve the overall GCN effectiveness, contributing for improvements in performance. This is most evident in the ''ST-GCN + REGINA'' configuration, where an improvement of 2.1 percentual points (i.e., decreasing the Top-1 error rates over 12\%) in the Cross-Subject evaluation benchmark. With respect to the 2s-AGCN network, the performance improvements were not as much significant, and were mostly observed in the Cross-View setting. In our viewpoint, this was due to the learnable graph topology of that technique, which has additional flexibility to better suit the hierarchical structure of GCNs, reducing the benefits due to the inclusion of the SSM-based handcrafted features.

\begin{table}[h]
\begin{center}
\begin{tabular}{|C{0.001\textwidth}|l{0.15\textwidth}|C{0.044\textwidth}|C{0.044\textwidth}|C{0.044\textwidth}|C{0.044\textwidth}|}
\cline{2-6}
\multicolumn{1}{c|}{} & 
\textbf{\footnotesize{Method}} & 
\textbf{\footnotesize{$\mathbf{Top\text{-}1}$}} &  \textbf{\footnotesize{$\mathbf{Top\text{-}5}$}}& \textbf{\footnotesize{$\mathbf{Top\text{-}1}$}} &  \textbf{\footnotesize{$\mathbf{Top\text{-}5}$}} \\
\cline{2-6}
\noalign{\smallskip}
\hline
\parbox[t]{1mm}{\multirow{5}{*}{\hspace{-0.21em}\rotatebox[origin=c]{90}{\footnotesize{\textbf{NTU RGB-D}}}}}
& \multicolumn{1}{c}{ \cellcolor{gray!25} }
& \multicolumn{2}{c}{ \cellcolor{gray!25}  \footnotesize{\emph{Cross-Subject} }}
& \multicolumn{2}{c|}{ \cellcolor{gray!25}  \footnotesize{\emph{Cross-View} }}\\
&\scriptsize{Js-AGCN~\cite{shi2019two}} & 
\scriptsize{$85.33$}  & 
\scriptsize{$97.04$}  &
\scriptsize{$93.72$}  & 
\scriptsize{$99.12$}  \\
&\scriptsize{\textbf{Js-AGCN} + \textbf{REGINA}} & 
\scriptsize{$\mathbf{86.62}$}  & 
\scriptsize{$\mathbf{97.51}$}  &
\scriptsize{$\mathbf{93.91}$}  & 
\scriptsize{$\mathbf{99.20}$}  \\
\cline{2-6}
&\scriptsize{Bs-AGCN~\cite{shi2019two}} & 
\scriptsize{$86.21$}  & 
\scriptsize{$97.09$}  &
\scriptsize{$93.18$}  & 
\scriptsize{$98.93$}  \\
&\scriptsize{\textbf{Bs-AGCN} + \textbf{REGINA}} & 
\scriptsize{$\mathbf{86.23}$}  & 
\scriptsize{$\mathbf{97.77}$}  &
\scriptsize{$\mathbf{93.32}$}  & 
\scriptsize{$\mathbf{99.41}$}  \\\hline

\end{tabular}
\vspace{-1em}
\end{center}
\caption{Comparison of \textbf{REGINA effectiveness in Multi-stream architectures}. Performance comparison regarding the integration of  REGINA in different feature streams (joints and bones) over both benchmarks of the NTU RGB+D \cite{shahroudy2016ntu} set.}
\label{tab:ablation_stream}
\end{table}

\textbf{Multi-stream architectures + REGINA fusion}. Most of the state-of-the-art methods utilize multi-stream fusion strategies. As a baseline to verify whther REGIna also contributes for improvements in performance for such kind of architectures, we considered a two-stream adaptive GCN (2s-AGCN) \cite{shi2019two}, which resorts not only to joints information (first-order) but also the second-order information (bones) of the skeleton-based data. Each stream of this framework is trained individually, combining both features to obtain the final class prediction.

The bones information is found by the difference between coordinates of two connected joints. For instance, for a bone with its source joint $\bm{v}_i = (x_i,y_i,z_i)$ and  target joint $\bm{v}_j = (x_j,y_j,z_j)$, the corresponding vector is given by $\bm{b}_{\bm{v}_i, \bm{v}_j} = (x_j-x_i,y_j-y_i,z_j-z_i)$. Considering that the human skeleton is an acyclic graph, we can assign each bone to a unique target joint. Moreover, as the skeleton central joint cannot be assigned to any bone, there is one bone less than joins in a skeleton, and - thus - an \emph{empty bone} was added to the central joint to simplify the network design. We then obtain the self-similarity descriptor for the second-order information,  similarly to the joints graph described in \eqref{eq:1}.

\begin{table*}[h]
\begin{center}
\begin{tabular}{|C{0.001\textwidth}|l{0.128\textwidth}|C{0.13\textwidth}|C{0.057\textwidth}|C{0.057\textwidth}|C{0.057\textwidth}|C{0.057\textwidth}|}
\cline{2-7}
\multicolumn{1}{c|}{} & 
\textbf{\footnotesize{Method}} & 
\textbf{\footnotesize{Type}} & 
\textbf{\footnotesize{$\mathbf{Top\text{-}1}$}} &  \textbf{\footnotesize{$\mathbf{Top\text{-}5}$}}& \textbf{\footnotesize{$\mathbf{Top\text{-}1}$}} &  \textbf{\footnotesize{$\mathbf{Top\text{-}5}$}} \\
\cline{2-7}
\noalign{\smallskip}
\hline
\parbox[t]{1mm}{\multirow{15}{*}{\hspace{-0.21em}\rotatebox[origin=c]{90}{\footnotesize{\textbf{NTU RGB-D}}}}}
& \multicolumn{2}{c}{ \cellcolor{gray!25} }
& \multicolumn{2}{c}{ \cellcolor{gray!25}  \footnotesize{\emph{Cross-Subject} }}
& \multicolumn{2}{c|}{ \cellcolor{gray!25}  \footnotesize{\emph{Cross-View} }}\\
&\scriptsize{Lie Group~\cite{vemulapalli2014human}} & 
\scriptsize{Handcrafted}  &
\scriptsize{$50.1$}  &
\scriptsize{$-$}  &
\scriptsize{$52.8$}  &
\scriptsize{$-$}  \\
\cline{2-7}
&\scriptsize{Deep LSTM~\cite{shahroudy2016ntu}} & 
\scriptsize{RNN}  &
\scriptsize{$60.7$}  & 
\scriptsize{$-$}  &
\scriptsize{$67.3$}  & 
\scriptsize{$-$}  \\
&\scriptsize{PA-LSTM~\cite{shahroudy2016ntu}} & 
\scriptsize{RNN}  &
\scriptsize{$62.9$}  & 
\scriptsize{$-$}  &
\scriptsize{$70.3$}  & 
\scriptsize{$-$}  \\
&\scriptsize{2s 3DCNN~\cite{liu2017two}} & 
\scriptsize{CNN}  &
\scriptsize{$66.8$}  & 
\scriptsize{$-$}  &
\scriptsize{$72.6$}  & 
\scriptsize{$-$}  \\
&\scriptsize{ST-LSTM~\cite{liu2016spatio}} & 
\scriptsize{RNN}  &
\scriptsize{$69.2$}  & 
\scriptsize{$-$}  &
\scriptsize{$77.7$}  & 
\scriptsize{$-$}  \\
&\scriptsize{TCN~\cite{soo2017interpretable}} & 
\scriptsize{CNN}  &
\scriptsize{$74.3$} & 
\scriptsize{$-$} &
\scriptsize{$83.1$}  & 
\scriptsize{$-$}  \\
&\scriptsize{VA-LSTM~\cite{zhang2017view}} & 
\scriptsize{RNN}  &
\scriptsize{$79.4$} & 
\scriptsize{$-$} &
\scriptsize{$87.6$}  & 
\scriptsize{$-$}  \\
&\scriptsize{Syn-CNN~\cite{liu2017enhanced}} & 
\scriptsize{CNN}  &
\scriptsize{$80.0$} & 
\scriptsize{$-$} &
\scriptsize{$87.2$}  & 
\scriptsize{$-$}  \\
\cline{2-7}
&\scriptsize{ST-GCN~\cite{yan2018spatial}} & 
\scriptsize{Graph}  &
\scriptsize{$81.51$}  & 
\scriptsize{$96.92$}  &
\scriptsize{$88.34$}  & 
\scriptsize{$98.33$}  \\
&\scriptsize{ST-GCN + REGINA} & 
\scriptsize{Graph +  Handcrafted}  &
\scriptsize{$\mathbf{83.62}$}  & 
\scriptsize{$\mathbf{97.71}$}  & 
\scriptsize{$\mathbf{90.12}$}  & 
\scriptsize{$\mathbf{99.00}$}  \\
\cline{2-7}
&\scriptsize{DPRL+GCNN~\cite{tang2018deep}} & 
\scriptsize{Graph}  &
\scriptsize{$83.5$}  & 
\scriptsize{$-$}  &
\scriptsize{$89.8$}  & 
\scriptsize{$-$}  \\
&\scriptsize{AS-GCN~\cite{li2019actional}} & 
\scriptsize{Graph}  &
\scriptsize{$86.8$}  & 
\scriptsize{$-$}  &
\scriptsize{$94.2$}  & 
\scriptsize{$-$}  \\
\cline{2-7}
&\scriptsize{2s-AGCN~\cite{shi2019two}} & 
\scriptsize{Graph}  &
\scriptsize{$\mathbf{88.51}$}  & 
\scriptsize{$\mathbf{98.43}$}  &
\scriptsize{$95.12$}  & 
\scriptsize{$99.14$}  \\
&\scriptsize{2s-AGCN + REGINA} &
\scriptsize{Graph + Handcrafted}  &
\scriptsize{$88.44$}  & 
\scriptsize{$98.22$}  &
\scriptsize{$\mathbf{95.21}$}  & 
\scriptsize{$\mathbf{99.38}$}  \\\hline

\end{tabular}
\vspace{-1em}
\end{center}
\caption{\textbf{NTU RGB+D recognition results}. Performance summary of the  REGINA architecture fused to the ST-GCN and 2s-AGCN architectures, with respect to the state-of-the-art methods in skeleton-based action recognition over both benchmarks from the NTU RGB+D \cite{shahroudy2016ntu} dataset.}
\label{tab:sota}
\end{table*}

The results are shown in Table \ref{tab:ablation_stream}. As can be seen,  the REGINA scheme has - again - contributed for improvements in the results. This is more evident in the joint-stream than in the bone-stream, which was justified by the fact that joints are less informative and discriminative than bones, where we already have the length and direction information.

\subsection{State-of-the-art Comparison}
\label{sec:sota}

To objectively perceive whether REGINA provides consistent advances over the state-of-the-art, in this section we compare the results attained by REGINA with/without the ST-GCN and 2s-AGCN architectures in both benchmarks of the NTU RGB+D \cite{shahroudy2016ntu} dataset. As previously stated, the state-of-the-art can be considered to comprise three categories: 1) handcrafted feature methods; 2) manually structured skeleton data approaches (CNNs and RNNs); and 3) and graph-based architectures. Our results are summarized in Table \ref{tab:sota}, being the methods grouped according to this criterion, in the ''Type'' column. As can be seen, REGINA is the first attempt to combine methods of two of out of these families: handcrafted features-based and graph-based networks.

The first observation was the overwhelming performance of graph-based architectures, when compared to the manually structured-based skeleton ones, such as CNNs and RNNs. Even though the latter approaches can be considered to still successfully extract structural information according to their data-driven architectures, it was evident solid improvements can be attained when modeling the skeleton data as a graph. Furthermore, despite the relatively poor performance of standalone handcrafted-based shallow representations, an overall improvement in performance can be obtained if handcrafted-based features are seemingly integrated more sophisticated Graph-based techniques, as the results of ''ST-GCN + REGINA'' (for both \emph{Cross-Subject} and \emph{Cross-View} settings) and ''2s-AGCN + REGINA'' (for the \emph{Cross-View} setting).

The results obtained also turn evident the relatively poor performance of the ST-GCN model, when compared to the deep reinforcement learning GCN (DPRL+GCNN) \cite{tang2018deep}, where this method resorts to reinforcement learning to select keyframes through a frame distillation network to refine the GCN. However, the ST-GCN enhanced by our proposal is able to overcome DPRL-GCNN performance, which we also consider to provide a sifgnificant cue over the improvements that would also be attained if fused to other similar techniques. In our view, this is mostly due to the fact that the optimized temporal adjacency matrix provides self-attention information to specific frames, while keeping the property of automated training in an end-to-end manner. Overall, our experiments point out that using graph convolutional networks fused to handcrafted-based feature layer can be considered an improvement for action recognition. Specifically,   REGINA is able to achieve competitive results, bringing gains up to 2.1 and 1.3 percentile points within the ST-GCN and 2s-AGCN, respectively, and overcoming the  state-of-the-art methods on the NTU RGB+D \cite{shahroudy2016ntu} dataset.

\subsection{Ablation Studies}
\label{sec:ablation}
In order to perceive the most important REGINA components and evaluate the overall robustness of our method to changes in parameterization, we varied the distance metrics used to obtain the self-similarity descriptor, the convolutional kernel size used in the handcrafted-feature fusion. Also, we performed additional experiments that show the importance of the learnable temporal adjacency importance over the handcrafted features.

\newcolumntype{C}[1]{>{\centering\arraybackslash}p{#1}}
\newcolumntype{l}[1]{>{\arraybackslash}p{#1}}

\begin{table}[h]
\begin{center}
\begin{tabular}{|C{0.001\textwidth}|l{0.17\textwidth}|C{0.044\textwidth}|C{0.044\textwidth}|}
\cline{2-4}
\multicolumn{1}{c|}{} & 
\textbf{\footnotesize{Method}} & 
\textbf{\footnotesize{$\mathbf{Top\text{-}1}$}} &  \textbf{\footnotesize{$\mathbf{Top\text{-}5}$}} \\
\cline{2-4}
\noalign{\smallskip}
\hline
\parbox[t]{1mm}{\multirow{4}{*}{\hspace{-0.21em}\rotatebox[origin=c]{90}{\footnotesize{\textbf{NTU RGB-D}}}}}
& \multicolumn{1}{c}{ \cellcolor{gray!25} }
& \multicolumn{2}{c|}{ \cellcolor{gray!25}  \footnotesize{\emph{Cross-View} }}\\
&\scriptsize{ST-GCN~\cite{yan2018spatial}} & 
\scriptsize{$88.34$}  & 
\scriptsize{$98.33$}  \\
&\scriptsize{ST-GCN + REGINA $(\ell-1)$}  & 
\scriptsize{$89.33$}  & 
\scriptsize{$98.62$}  \\
&\scriptsize{\textbf{ST-GCN} + \textbf{REGINA} $(\bm{\ell-2})$} & 
\scriptsize{$\mathbf{90.12}$}  & 
\scriptsize{$\mathbf{99.00}$}  \\\hline

\end{tabular}
\vspace{-1em}
\end{center}
\caption{\textbf{Self-similarity's distance metrics results}. Performance summary of different distance metrics over the Cross-View benchmark of NTU RGB+D \cite{shahroudy2016ntu}.}
\label{tab:ablation_dist}
\end{table}

\textbf{Role of the distance metrics}. We compare two different distance metrics between time-frame pairs in obtaining each skeleton sequence self-similarity matrix: a) the Manhattan distance, also known as $\ell-1$ norm, which expresses the sum of individual coordinates shifting difference between two skeleton poses from different temporal instances; and b) the Euclidean distance, also known as $\ell-2$ norm, that expresses the direct distance from one pose to another from the temporal skeleton difference perspective. As shown in Table \ref{tab:ablation_dist}, the self-similarity descriptor using with the $\ell-1$ norm and fused to the ST-GCN \cite{yan2018spatial} network is already able to improve the baseline performance. However, since the similarity between two time-frame pairs only considers the distance between the same joints, without any structural information, directions or angles, better expressiveness of the temporal kinematics can be captured by the $\ell-2$ norm, which justifies the optimal performance observed for that distance function.

\begin{table}[h]
\begin{center}
\begin{tabular}{|C{0.001\textwidth}|l{0.22\textwidth}|C{0.044\textwidth}|C{0.044\textwidth}|}
\cline{2-4}
\multicolumn{1}{c|}{} & 
\textbf{\footnotesize{Method}} & 
\textbf{\footnotesize{$\mathbf{Top\text{-}1}$}} &  \textbf{\footnotesize{$\mathbf{Top\text{-}5}$}} \\
\cline{2-4}
\noalign{\smallskip}
\hline
\parbox[t]{1mm}{\multirow{4}{*}{\hspace{-0.21em}\rotatebox[origin=c]{90}{\footnotesize{\textbf{NTU RGB-D}}}}}
& \multicolumn{1}{c}{ \cellcolor{gray!25} }
& \multicolumn{2}{c|}{ \cellcolor{gray!25}  \footnotesize{\emph{Cross-View} }}\\
&\scriptsize{ST-GCN~\cite{yan2018spatial}} & 
\scriptsize{$88.34$}  & 
\scriptsize{$98.33$}  \\
&\scriptsize{ST-GCN + REGINA (w/o Conv-Layer)} & 
\scriptsize{$88.62$}  & 
\scriptsize{$98.43$}  \\
&\scriptsize{\textbf{ST-GCN} + \textbf{REGINA (w Conv-Layer)}} & 
\scriptsize{$\mathbf{90.12}$}  & 
\scriptsize{$\mathbf{99.00}$}  \\\hline

\end{tabular}
\vspace{-1em}
\end{center}
\caption{\textbf{Effectiveness of the temporal adjacency matrix}. Performance comparison of the results obtained with/without the convolutional layer. Results are given for the \emph{Cross-View} benchmark of NTU RGB+D \cite{shahroudy2016ntu} set.}
\label{tab:ablation_conv}
\end{table}

\textbf{Learnable temporal adjacency matrix effectiveness}. As described in Section \ref{ssec:graph}, the temporal adjacency matrix consists of a convolutional layer on top of the handcrafted feature, which then becomes a learnable matrix that controls the GCN temporal receptive fields through the self-similarity matrix. For ablation purposes, we manually removed the convolutional layer and observed the resulting REGINA performance, with results shown in Table \ref{tab:ablation_conv}. The obtained values show that the learnable temporal adjacency matrix has the ability to adapt to different semantical levels, which is beneficial for the hierarchical GCN as different layers contain multilevel semantic information. This way, REGINA architecture without the convolutional layer can be seen as a rigid and pre-defined temporal configuration (self-similarity descriptor) to every layer. Despite still contributing for improvements in performance than the ST-GCN \cite{yan2018spatial} alone, this fusion scheme is clearly sub-optimal. Consequently, the graph convolution architecture with the learnable temporal adjacency matrix achieves consistently higher performance than its counterpart.

\begin{table}[h]
\begin{center}
\begin{tabular}{|C{0.001\textwidth}|l{0.18\textwidth}|C{0.044\textwidth}|C{0.044\textwidth}|}
\cline{2-4}
\multicolumn{1}{c|}{} & 
\textbf{\footnotesize{Method}} & 
\textbf{\footnotesize{$\mathbf{Top\text{-}1}$}} &  \textbf{\footnotesize{$\mathbf{Top\text{-}5}$}} \\
\cline{2-4}
\noalign{\smallskip}
\hline
\parbox[t]{1mm}{\multirow{6}{*}{\hspace{-0.21em}\rotatebox[origin=c]{90}{\footnotesize{\textbf{NTU RGB-D}}}}}
& \multicolumn{1}{c}{ \cellcolor{gray!25} }
& \multicolumn{2}{c|}{ \cellcolor{gray!25}  \footnotesize{\emph{Cross-View} }}\\
&\scriptsize{ST-GCN + REGINA ($k = 1$)} & 
\scriptsize{$88.93$}  & 
\scriptsize{$98.61$}  \\
&\scriptsize{\textbf{ST-GCN} + \textbf{REGINA} ($\mathbf{k = 3}$)} & 
\scriptsize{$\mathbf{90.12}$}  & 
\scriptsize{$\mathbf{99.00}$}  \\
&\scriptsize{ST-GCN + REGINA ($k = 5$)} & 
\scriptsize{$89.31$}  & 
\scriptsize{$98.63$}  \\
&\scriptsize{ST-GCN + REGINA ($k = 7$)} & 
\scriptsize{$89.12$}  & 
\scriptsize{$98.51$}  \\
&\scriptsize{ST-GCN + REGINA ($k = 9$)} & 
\scriptsize{$89.10$}  & 
\scriptsize{$98.50$}  \\\hline

\end{tabular}
\vspace{-1em}
\end{center}
\caption{\textbf{Importance of the kernel size} from the learnable temporal adjacency matrix over the handcrafted feature. Results regard the recognition performance over the \emph{Cross-View} benchmark of the NTU RGB+D \cite{shahroudy2016ntu} set, with different kernel sizes ($k$) of the convolutional layer on top of the handcrafted feature.}
\label{tab:ablation_k}
\end{table}


\textbf{Exploring the convolutional kernel size}. Finally, we conducted some experiments on REGINA, when varying the kernel size ($k$) of the learnable adjacency matrix (\textit{Conv2D Layer} in Fig. \ref{fig:regina_unit}). Based in the results reported in Table \ref{tab:ablation_k}, we can see that our model performance reaches a peak when $k=3$, meaning that we have nine temporal comparisons (from the self-similarity matrix) being fed into the convolutional kernel. For smaller kernel sizes ($k=1$), we only perform one comparison per convolution  operation, which decreases the information feed to the subsequent layer of the network. In opposition, for larger $k>3$ values, the performance was observed to decrease again, which we justified to the too high number of temporal comparisons fed into the convolutional kernel, that affects the discriminability of the descriptor itself (i.e., it is a weighted mean of too many values). This has a negative effect in the extraction of short-term temporal adjacency features, which results in lower performance.

 \section{Conclusions and Further Work}
 \label{sec:Conclusions}
  The automated and reliable recognition of human actions has been the focus of substantial research efforts, being accepted that it will support various applications in the security/forensics domains. 
  In this context, Graph Convolutional Networks-based (GCNs) approaches have been advancing consistently the state-of-the-art performance, being currently the most popular kind of approaches for this problem. In this paper, we proposed REGINA, an architecture that can be faithfully integrated into the state-of-the-art GCNs techniques and enables consistent improvements in performance over the baselines. The idea is to fuse the information naturally extracted by GCNs to \emph{particularly interesting} handcrafted representations (SSM), which provide complementary information to the networks and intuitively enables that the networks focus in extracting alternate (orthogonal) information to the SSM during the optimization process. As a result, the network has access to additional cues that were observed to contribute for improvements in performance. Also, the proposed REGINA scheme is able to overcome the limitations of conventional temporal convolutions, by capturing discriminative features over the temporal dynamics. Our experiments pointed out for highly encouraging results, where REGINA + state-of-the-art GCN techniques - is able to push forward their performance on two challenging benchmarks. Our efforts are currently focused in finding/describing automatically other kinds of handcrafted features that will also be positively combined to the automatically extracted by GCN-based networks. 


%



\section*{Acknowledgments}
This work is funded by FCT/MEC through national funds and co-funded by FEDER - PT2020 partnership agreement under the project UIDB//50008/2020. Also, it was supported by operation Centro-01-0145-FEDER-000019 - C4 - Centro de Compet\^{e}ncias em Cloud Computing, co-funded by the European Regional Development Fund (ERDF) through the Programa Operacional Regional do Centro (Centro 2020), in the scope of the Sistema de Apoio \`{a} Investiga\c{c}\~{a}o Cient\'{i}fica e Tecnol\'{o}gica - Programas Integrados de IC\&DT, and supported by `FCT - Fundação para a Ciência e Tecnologia' through the research grants `2020.04588.BD' and `UI/BD/150765/2020'.

\ifCLASSOPTIONcaptionsoff
  \newpage
\fi

\input{ms.bbl}

%









\end{document}